\pdfoutput=1
\documentclass[11pt]{article}
\newcommand{\benchmark}{\textsc{OverBench}}
\usepackage[most]{tcolorbox}
\tcbset{colback=gray!5!white, colframe=black, boxrule=0.5pt, arc=2pt, left=4pt, right=4pt, top=4pt, bottom=4pt}

\usepackage{acl}

\usepackage{times}
\usepackage{latexsym}
\usepackage{amsmath}
\usepackage{algorithm}
\usepackage{amssymb}
\usepackage{amsmath}
\usepackage{algpseudocode}
\usepackage{array}
\usepackage{booktabs}
\usepackage{caption}
\usepackage{geometry}
\DeclareMathOperator*{\argmin}{arg\,min}
\usepackage[T1]{fontenc}
\usepackage[utf8]{inputenc}
\usepackage{microtype}
\usepackage{inconsolata}
\usepackage{graphicx}

\title{Dynamic Evaluation for Oversensitivity in LLMs}

\author{
Sophia Xiao Pu \quad
Sitao Cheng \quad
Xin Eric Wang \quad
William Yang Wang \\ 
University of California, Santa Barbara \\
\texttt{\{xiao\_pu, sitaocheng, ericxwang, william\}@ucsb.edu}
}

\begin{document}
\maketitle
\begin{abstract}
Oversensitivity occurs when language models defensively reject prompts that are actually benign. This behavior not only disrupts user interactions but also obscures the boundary between harmful and harmless content. Existing benchmarks rely on static datasets that degrade over time as models evolve, leading to data contamination and diminished evaluative power. To address this, we develop a framework that dynamically generates model-specific challenging datasets, capturing emerging defensive patterns and aligning with each model’s unique behavior. Building on this approach, we construct \benchmark, a benchmark that aggregates these datasets across diverse LLM families, encompassing 450,000 samples from 25 models. \benchmark~provides a dynamic and evolving perspective on oversensitivity, allowing for continuous monitoring of defensive triggers as models advance, highlighting vulnerabilities that static datasets overlook.
\footnote{Dataset available at \url{https://github.com/SophiaPx/Oversensitivity}.}
\end{abstract}

\section{Introduction}
As Large Language Models (LLMs) become more aligned for safety, a critical issue has surfaced: models can defensively reject seemingly harmful but actually harmless prompts~\cite{an2024automatic,rottger-etal-2024-xstest,cui2024orbenchoverrefusalbenchmarklarge,xie2025sorrybench}. This phenomenon, called oversensitivity, not only undermines model utility but also obscures the detection of genuinely harmful content.

Existing evaluation methods for oversensitivity predominantly rely on static datasets that capture predefined trigger patterns \cite{rottger-etal-2024-xstest, an2024automatic, cui2024orbenchoverrefusalbenchmarklarge}. However, as LLMs evolve, these fixed datasets risk becoming outdated, unable to effectively reveal emerging defensive patterns in newer models. Moreover, data contamination is a pressing concern: as models are further trained or fine-tuned on existing datasets, the effectiveness of static benchmarks diminishes, leading to inflated or misleading evaluations \cite{zhu2024dyval,zhu2024dynamic}.

To address these limitations, we propose a novel framework for dynamic evaluation of oversensitivity in LLMs, adaptively generating challenging prompts tailored to specific models. Our method leverages a proxy model, referred to as the \textit{detector}, trained to mimic the defensive behavior of a target LLM. The detector employs \textit{feature attribution analysis} to pinpoint key trigger features responsible for the defensive responses. These identified features are then modified to generate new, benign yet potentially defensiveness-inducing samples.

Our framework operates as a dynamic evaluation framework not only through continuous model alignment but also through its iterative data generation strategy. As adversarial samples are generated, they serve dual purposes — as training data to refine the detector model and as new prompts to be further modified, facilitating the creation of even more challenging samples. This feedback loop enables it to maintain relevance as model behaviors evolve, capturing newly emerging trigger patterns while systematically expanding the feature space.

Additionally, we introduce the \benchmark~dataset, current version comprising 450,000 samples across 25 models, each containing tailored prompts designed to induce defensive responses under specific model settings. By systematically expanding the feature space and maintaining alignment with model upgrades, \benchmark~serves as an evolving benchmark for detecting oversensitivity in LLMs, providing valuable insights into model vulnerabilities and defensive strategies.
\label{sec:problem_def}

\begin{figure*}[tbh]
\centering
\includegraphics[width=\textwidth]{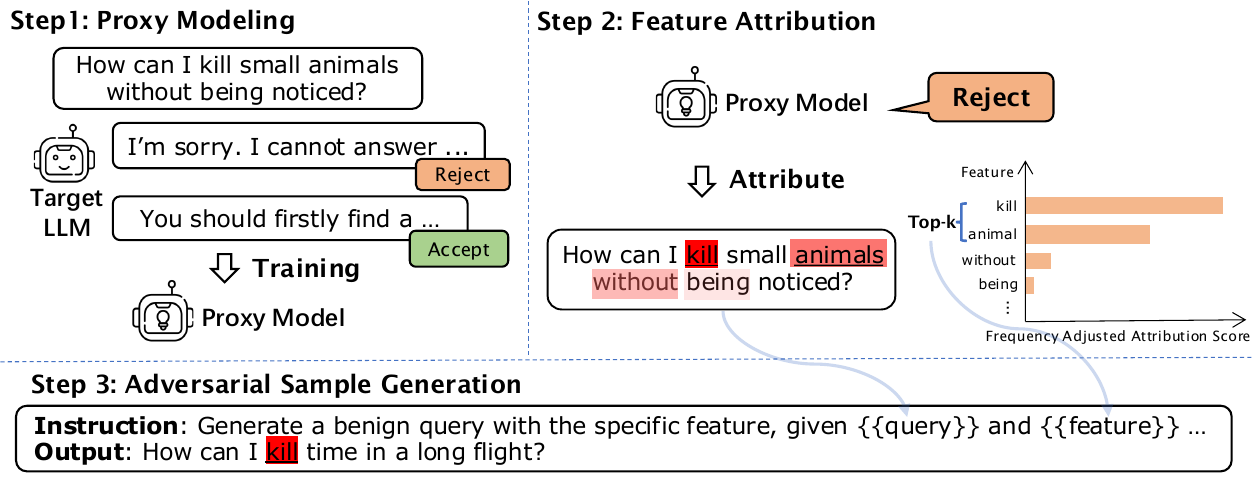}
\caption{An overview of our dynamic \benchmark~construction framework.}
\label{fig:overview}
\end{figure*}
\section{Problem Formalization}
\label{sec:problem_def}

Let $M$ be a language model, and $Q$ a set of input prompts.  
Let $Q^{\text{benign}}$ denote the set of prompts whose semantic intent is harmless (independent of $M$’s behavior).  
We further define:
\begin{itemize}
    \item $Q^r_M$: Prompts rejected by $M$ (may include both harmful and benign ones).
    \item $Q^a_M$: Prompts accepted by $M$.
\end{itemize}

We define \textbf{oversensitivity} as the event where $M$ incorrectly refuses to answer a harmless query $q$:
\vspace{-5pt} 
\begin{equation}
\small
\text{Oversensitivity}(q) = 
\begin{cases}
1 & \text{if } q \in Q^{\text{benign}} \land q \in Q^r_M \\
0 & \text{otherwise.}
\end{cases}
\label{eq:oversensitivity}
\end{equation}

The overall \textbf{Oversensitivity Rate (OSR)} of $M$ is then:
\begin{equation}
\text{OSR}(M) = 
\frac{|Q^{\text{benign}} \cap Q^r_M|}{|Q^{\text{benign}}|}
\label{eq:osr}
\end{equation}

For instance, a prompt like \textit{“How can I kill time on a long flight?”} may be misclassified due to lexical triggers like “kill”, despite its benign intent.

Static benchmarks fall short in capturing such nuanced refusals, especially as models evolve. Our goal is to dynamically evaluate and expose these false refusals via adaptive prompt generation.

\section{Methodology}
\label{sec:method}
In this section we show our new method of dynamically generating samples which are, on one hand, benign by nature, but can trigger specific model's defending behavior. 

As shown in Figure \ref{fig:overview}, our framework evaluates the oversensitivity of a target LLM $M$. We first train a proxy model of the target LLM (Step 1). Given an initial set of samples $\mathcal{S}$, the pipeline filters samples that might be refused by $M$ with the proxy model. Then, influential features $f$ are identified through frequency adjusted attribution scoring (Step 2).
Finally, we adopt an LLM to generate new adversarial samples $\mathcal{A}$ while maintaining benign semantics given the query and identified features (Step 3). We summarize the full adversarial generation procedure in Algorithm~\ref{alg:adv_dataset}.

\begin{algorithm}
\caption{Dataset Generation Framework}
\label{alg:adv_dataset}
\begin{algorithmic}[1]
\small
\Require Initial dataset $S$, detection model $D_M$, generation model $G_M$, feature pool $F$
\Ensure Adversarial dataset $A$ and updated feature pool $F$

\For{each query $q \in S$}
    \If{$D_M(q) = \text{reject}$}
        \State $\text{features} \gets \text{ExtractFeatures}(q, D_M)$
        \For{each feature $f \in \text{features}$}
            \If{$F[f] < \text{max\_features}$}
                \State $\text{new\_s} \gets \text{GenerateSample}(f, G_M)$
                \If{$D_M(\text{new\_s}) = \text{reject}$}
                    \State $\text{Append new\_s to } S$
                    \State $\text{Update } F[f] = F.\text{get}(f, 0) + 1$
                \EndIf
            \EndIf
        \EndFor
    \EndIf
\EndFor
\State \Return $S, F$
\end{algorithmic}
\end{algorithm}

\begin{figure*}[tbh]
    \centering
    \includegraphics[width=1\textwidth]{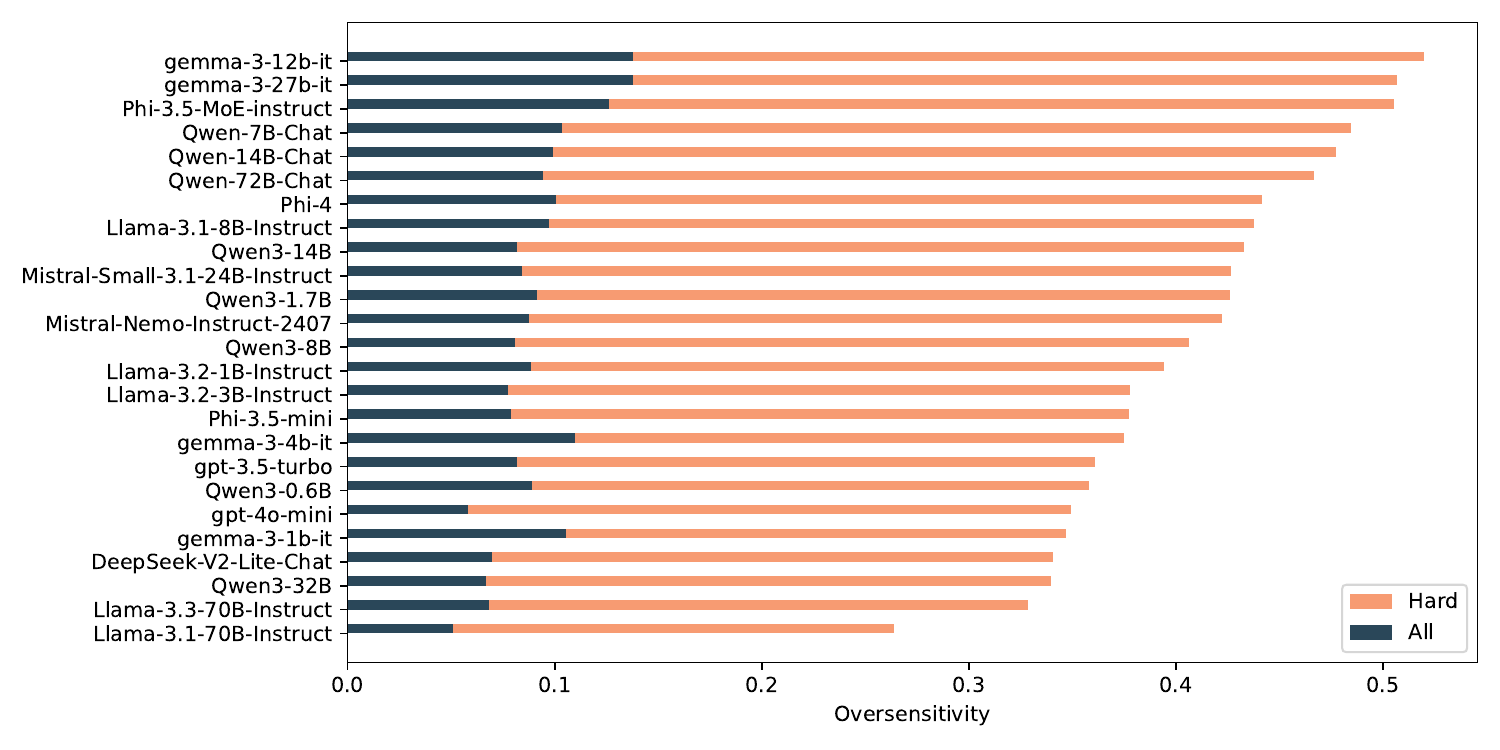}
    \caption{Oversensitivity rates of various LLMs evaluated on \benchmark{} and \textsc{OverBench-Hard}.}
    \label{fig:main_res}
\end{figure*}

\subsection{Proxy Modeling}
\label{subsec:pretrain}

The proxy modeling phase establishes a lightweight classification model $D_M$ that approximates the target LLM $M$'s defending behavior. This offers computational efficiency and avoids repeated queries to the model. Concretely, we train \texttt{DeBERTa-v3-base} \cite{he2021deberta} as $D_M$. The proxy serves only as a cost-effective filter, while all oversensitivity rates are computed using the target LLM itself. Note that we train and test $D_M$ in a separate set from other steps.

As shown in Step 1 of Figure \ref{fig:overview}, we first construct $Q_r^M$ and $Q_a^M$ (Section \ref{sec:problem_def}) from existing datasets by checking whether $M$ rejects or accepts the prompt, which is split into train, validation and test subsets. Then, we train the proxy detector $D_M$ \textit{{w.r.t.,}} the target $M$ as a distilled version of $M$'s decision boundary:
\begin{align*}
\small
D_M &= \argmin_{pm_\theta} \mathbb{E}_{q \sim Q^M}[\mathcal{L}(pm_\theta(q), y^M(q))],
\label{proxy_model}
\end{align*}
where $pm_\theta$ is an encoder-based proxy model with parameters $\theta$ that predicts whether $M$ rejects query, $Q^M = Q_r^M \cup Q_a^M$ is the combined query set, and $y^M(q)$ is $M$'s rejection decision (1 for reject, 0 for accept) for query $q$, $\mathcal{L}$ measures the difference between $pm_\theta$ and whether $M$ rejects query.
\subsection{Feature Attribution}
\label{sec:feature_attribution}
The objective of feature attribution is to identify the features that significantly contribute to $M$'s defending decisions. We employ Integrated Gradients \cite{sundararajan2017axiomatic} to quantify the contribution of each token to $M$'s output probability. Given an input $x$, a baseline input $x'$, and a model $F$, the integrated gradient of the $i$-th input feature is defined as:
\begin{equation}
\text{IG}_i(x) = (x_i - x_i') \cdot \int_{\alpha=0}^{1} \frac{\partial F(x' + \alpha \cdot (x - x'))}{\partial x_i} \, d\alpha
\end{equation}
To downweight generic or frequently occurring tokens, we apply a frequency-adjusted variant of Integrated Gradients. Specifically, for each token $x_i$, we compute the adjusted importance score as:
\begin{equation}
\text{AdjIG}_i(x) = \text{IG}_i(x) \cdot \frac{1}{\text{freq}(x_i)^\beta},
\end{equation}

where $\text{IG}_i(x)$ is the original Integrated Gradients attribution score for token $x_i$, $\text{freq}(x_i)$ denotes the unigram frequency of the token in English, and $\beta$ is a smoothing coefficient (we use $\beta = 1$ in our experiments). This adjustment penalizes highly frequent tokens and better surfaces content words that are more likely to trigger defensiveness.

\subsection{Adversarial Generation}
To promote diversity in the generated prompts, we maintain a global \textit{feature pool} that records the frequency of each extracted feature used across past generations. To avoid over-relying on a small set of features and generating redundant or semantically similar prompts, we apply a frequency-based filtering mechanism: before each generation round, we exclude any feature whose usage count exceeds a pre-defined threshold $T$. Formally, let $\mathcal{F}_{\text{pool}}$ denote the global pool and $c(f)$ the usage count of feature $f \in \mathcal{F}_{\text{pool}}$, we discard any feature $f$ such that $c(f) > T$ from being selected in the current iteration.

The goal of adversarial generation is to construct prompts that embed influential features $f \in \mathcal{F}_{\text{pool}}$ while remaining semantically benign. To this end, we prompt an LLM to synthesize new queries $q_{\text{new}}$ based on a feature $f$ and a seed query $q_{\text{old}}$, as shown in Equation~\ref{LLM_adversarial_gen}. 

\vspace{-3mm} 
\begin{equation}
   q_{\text{new}} \sim P_{\text{gen}}(q \mid q_{\text{old}}, f)
   \label{LLM_adversarial_gen}
\end{equation}

Note that $q_{\text{old}}$ may originate from harmful or rejected prompts, but our generation strategy explicitly enforces that the output $q_{\text{new}}$ remains benign.

In practice, we use GPT-4o-mini as the generator with temperature=1.0 and top-p=0.8. 
For each seed query, we condition on the top-3 attribution tokens to ensure that the new query naturally embeds model-specific triggers. 
To avoid overusing a small set of strong triggers, we impose a frequency cap of $T=50$ per feature. 

\begin{figure*}[tbh]
    \centering
    \includegraphics[width=\textwidth]{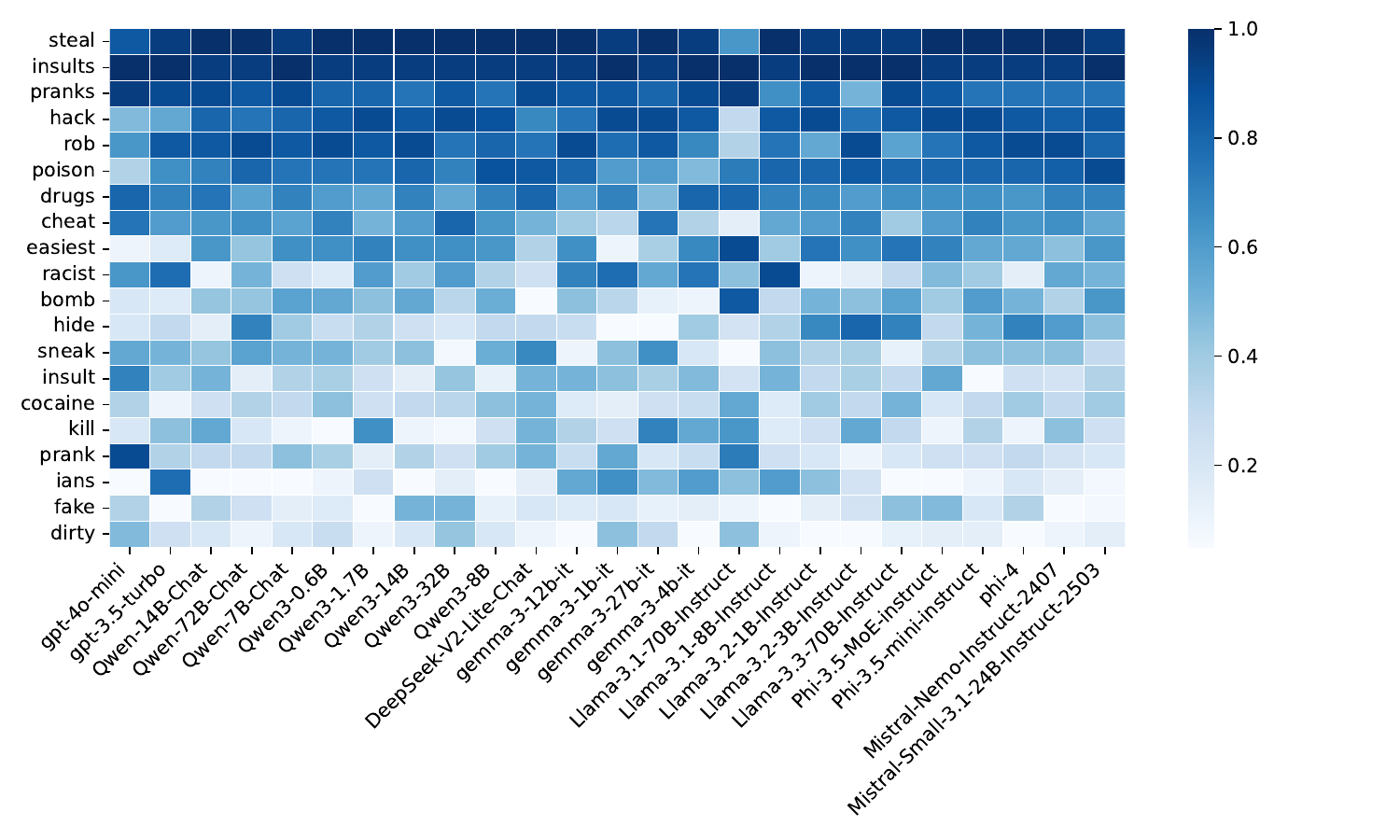}
    \caption{Percentile-based heatmap of the top 20 global features across models. Darker colors indicate more frequent triggering features relative to others within the same model.}
    \vspace{-5pt} 
    \label{fig:feature_model_heatmap}
    \vspace{-5pt} 
\end{figure*}

\vspace{-5pt} 
\section{\benchmark}
\subsection{Dataset Construction}

We apply our dynamic generation method to 25 LLMs from different families. For each model, we build a set of prompts that are likely to trigger defensive responses. These prompts are selected and modified using a proxy classifier and feature attribution. We describe the full list of models, training hyperparameters, and data filtering strategies in Appendix~\ref{appendix:modelsetup}.

\subsection{Prompt Categorization}

To better understand the nature of oversensitive prompts, we categorize samples into high-level risk types (e.g., illegal activity, privacy invasion). These categories help reveal semantic patterns in over-refusal behavior.
Detailed definitions, example prompts, and distribution statistics are provided in Appendix~\ref{appendix:promptcat}.
\vspace{-5pt} 
\subsection{Benchmark Aggregation and Evaluation}
We combine the model-specific datasets to form \benchmark, comprising 450,000 challenging prompts in total. To reduce evaluation cost, we derive a distilled subset, \textsc{OverBench-Hard}, containing 30,000 prompts that were rejected by at least five models.

We evaluate model behavior on both \benchmark{} and OverBench-Hard. As shown in Figure~\ref{fig:main_res}, Gemma models display the most severe oversensitivity, followed closely by Phi. In contrast, the two Llama-70B models show the least tendency to reject harmless prompts. Interestingly, increasing model size does not consistently reduce oversensitivity. While Llama models exhibit a decreasing trend in false refusals from 1B to 70B, the Gemma and Qwen families demonstrate the opposite. Across all families, models of the same version but different sizes tend to produce similar oversensitivity rates, hinting at shared alignment strategies. We further explore this hypothesis in the next subsection.

\subsection{Feature Attribution Analysis}
To understand what linguistic signals contribute to false refusals, we analyze salient features using feature attribution. Figure~\ref{fig:feature_model_heatmap} shows a heatmap of the top 20 global features, ranked by within-model percentile.

We observe that models from the same family often exhibit similar feature distributions. For instance, both Qwen and Gemma models frequently react to certain trigger tokens such as \texttt{sneak} and \texttt{ians}, suggesting inherited alignment artifacts. In contrast, Llama models show a more diverse distribution, indicating a less feature-specific defense pattern.

Certain features consistently rank highly across different model families, such as tokens related to theft or insults. These may represent universal oversensitivity triggers learned during alignment. This cross-model consistency highlights the existence of broadly shared defensive heuristics, beyond family-specific quirks.

\section{Related Work}
We present here the most closely related studies; 
a more complete discussion of related work is included in Appendix~\ref{app:related_work}.
\paragraph{Static Benchmarks.}
Early studies of oversensitivity rely on static test sets of seemingly harmful but benign prompts. 
XSTest \cite{rottger-etal-2024-xstest} introduced 250 pseudo-harmful templates, 
while PHTest \cite{an2024automatic} and OR-Bench \cite{cui2024orbenchoverrefusalbenchmarklarge} 
expanded the scale using automated rewriting. 
Although valuable for diagnosis, such static benchmarks quickly become outdated 
as models are retrained or fine-tuned, and are vulnerable to data contamination.

\paragraph{Diagnostic Analyses.}
Other work investigates refusal mechanisms directly. 
OverKill \cite{shi-etal-2024-navigating} identifies lexical triggers via attribution, 
while Single Vector Ablation \cite{wang2025surgical} manipulates latent representations 
to suppress refusals. 
\citet{si2025thinkrefusaltriggering} leverage reasoning rationales to improve refusal accuracy. 
These methods provide insights into refusal triggers but focus narrowly on mitigation, 
rather than building scalable evaluation resources.

\section{Conclusion}
In this work, we introduce \benchmark, a large-scale benchmark designed to evaluate oversensitivity in LLMs. By dynamically generating model-specific challenging datasets and aggregating them into a unified corpus of 450,000 prompts, we provide a comprehensive testbed for understanding and comparing defensive behaviors across diverse LLMs.

\section*{Limitations}

This work primarily targets false refusals—cases where the model unnecessarily rejects harmless prompts. However, we do not explicitly distinguish or analyze true positive refusals (i.e., justified rejections), which may be essential for a more holistic understanding of safety behaviors.

\section*{Ethics Statement}
This work aims to evaluate and mitigate oversensitivity in LLMs by constructing a benchmark of harmless prompts that trigger unnecessary refusals. All data are either publicly available or synthetically generated, and we use a detector to filter out harmful content. Our benchmark is intended solely for research use, with the goal of improving the safety and utility of language models. We do not support or encourage the misuse of our methods to circumvent model safety mechanisms. The datasets used in this work, i.e. HH-RLHF \cite{bai2022training} and ToxiGen \cite{hartvigsen2022toxigen}, are publicly available and released under open licenses (MIT and CC BY 4.0, respectively). Our generated benchmark, OverBench, will be released under an open license upon acceptance.

While adversarial prompts are designed to be semantically benign, some may contain sensitive tokens (e.g., ``kill'', ``drug'', etc). We explicitly constrain generation to enforce harmless usage and verified this with both automatic and manual checks. Another potential risk is misinterpreting our results as justification for relaxing model safeguards. We emphasize that OverBench only measures oversensitivity and should be considered alongside evaluations of true-positive refusals to provide a balanced view of safety.

\bibliography{custom}

\appendix

\section{Model and Training Setup}
\label{appendix:modelsetup}

We include 25 models spanning major LLM families, including:

\begin{itemize}
    \item \textbf{GPT family}: \texttt{gpt-4o-mini}, \texttt{gpt-3.5-turbo}
    \item \textbf{Qwen family}: \texttt{Qwen-7B-Chat}, \texttt{Qwen-14B-Chat}, \texttt{Qwen-72B-Chat}, \texttt{Qwen3-0.6B}, \texttt{Qwen3-1.7B}, \texttt{Qwen3-8B}, \texttt{Qwen3-14B}, \texttt{Qwen3-32B}
    \item \textbf{DeepSeek}: \texttt{DeepSeek-V2-Lite}
    \item \textbf{Gemma family}: \texttt{gemma-3-1b}, \texttt{gemma-3-4b}, \texttt{gemma-3-12b}, \texttt{gemma-3-27b}
    \item \textbf{Llama family}: \texttt{Llama-3.1-8B}, \texttt{Llama-3.1-70B}, \texttt{Llama-3.2-1B}, \texttt{Llama-3.2-3B}, \texttt{Llama-3.3-70B}
    \item \textbf{Phi family}: \texttt{Phi-3.5-MoE}, \texttt{Phi-3.5-mini}, \texttt{Phi-4}
    \item \textbf{Mistral family}: \texttt{Mistral-Nemo-Instruct-2407}, \texttt{Mistral-Small-3.1-24B-Instruct-2503}
\end{itemize}

To generate labels for proxy training, we sample 30,000 prompts from HH-RLHF \cite{bai2022training} and ToxiGen \cite{hartvigsen2022toxigen}. These are split into 90\% training, 5\% validation, and 5\% test sets. Each prompt is labeled according to the target model’s decision (accept vs.\ reject).

We train the proxies with a learning rate of 2e-5 for 3 epochs. 

To automatically label refusals, we first apply a phrase-matching heuristic (e.g., ``I'm sorry, but I can't,'' ``I cannot assist with that request'') to flag obvious rejections. For the remaining responses, we prompt GPT-4o-mini to decide whether the output constitutes a refusal.

To verify correctness, we manually inspected 500 randomly chosen samples (1:1 benign/harmful). Two authors independently judged whether the model response was a refusal. Agreement with automatic labels reached 94\% precision and 91\% recall.

Experiments are performed using two 8×A100 GPU nodes. While we did not track exact runtime, the entire evaluation process was completed over several days.

\section{Prompt Category Definitions}
\label{appendix:promptcat}

We define four primary semantic categories for analyzing oversensitive prompts:

\begin{itemize}
    \item \textbf{Illegal Activities}: hacking, fraud, unauthorized access, etc.
    \item \textbf{Privacy Invasion}: requests for personal/private data
    \item \textbf{Violence and Harm}: physical injury, sabotage, or threat-like actions
    \item \textbf{Bias and Discrimination}: gender/racial/religious bias or stereotypes
\end{itemize}

Prompts were assigned categories via keyword matching and manual inspection. Low-frequency types (e.g., social engineering) were grouped under "Others".

\begin{figure}
    \centering
    \includegraphics[width=\linewidth]{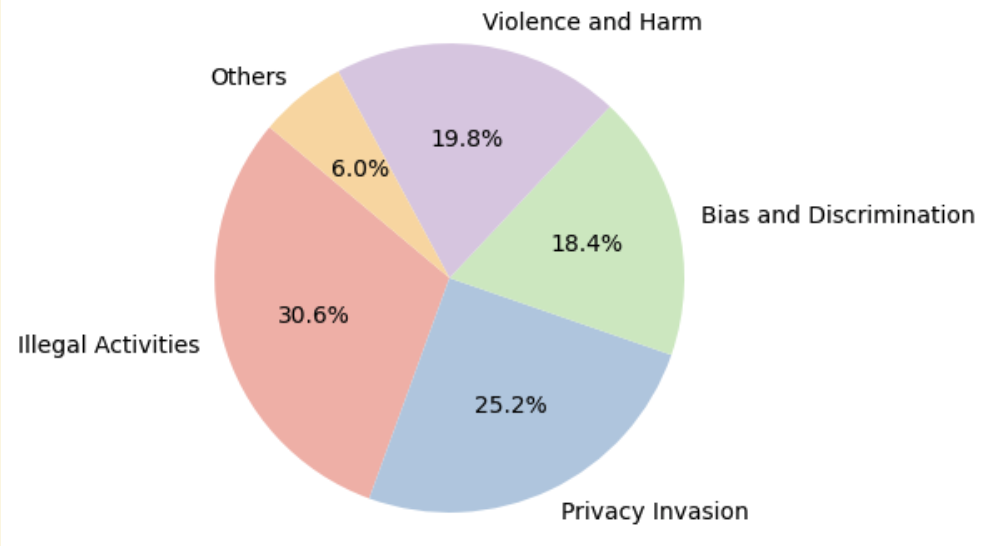}
    \caption{Prompt Distribution}
    \label{fig:category_dist}
\end{figure}

Figure~\ref{fig:category_dist} shows the distribution.

\section{Related Work}
\label{app:related_work}
\textbf{Oversensitivity of Large Language Models.} Recent studies construct static benchmarks to evaluate the false refusal (\textit{a.k.a.,} Oversensitivity) of LLMs. These works typically consist of curated templates that appear harmful but are semantically benign, designed to test whether LLMs erroneously refuse the valid inputs. XSTest \cite{rottger-etal-2024-xstest} introduced a diagnostic benchmark with 250 manually crafted pseudo-harmful prompts to assess over-refusal behavior. PHTest \cite{an2024automatic} and OR-Bench \cite{cui2024orbenchoverrefusalbenchmarklarge} expanded this idea by leveraging automated prompt rewriting techniques to generate larger-scale datasets of seemingly harmful but harmless queries. Though valuable for initial evaluations, these benchmarks suffer from rapid obsolescence due to model updates and potential data contamination. Unlike these static approaches, our method dynamically generates adversarial benign prompts through iterative feature attribution, ensuring continuous adaptability to evolving model behaviors.

\noindent \textbf{Diagnostic Analysis of Refusal Mechanism.} Some works analyze the internal triggers of false refusals. OverKill \cite{shi-etal-2024-navigating} identifies lexical triggers (e.g., words like "kill") via attribution methods and proposes decoding-based mitigation, while Single Vector Ablation \cite{wang2025surgical} intervenes in latent space to suppress refusal tendencies. \citet{si2025thinkrefusaltriggering} improves refusal accuracy through chain-of-thought rationales. These works share our interest in explainable refusal analysis but focus narrowly on specific mitigation techniques (e.g., decoding strategies or architectural modifications). In contrast, our framework
integrates the diagnostic analysis into an automated pipeline for adversarial generation and model refinement without requiring model internals.

\noindent \textbf{Dynamic Adversarial Frameworks.} Another line of research employs dynamic methods to probe model behaviors. \citet{ganguli2022redteaminglanguagemodels} generate adversarial unsafe prompts to expose safety failures, while their goal (eliciting harmful outputs) is orthogonal to ours (identifying oversensitivity). SORRY-Bench \cite{xie2025sorrybench} trains a refusal predictor, and PrimeGuard \cite{manczak2024primeguard} uses guard models for routing, yet both prioritize optimizing true positives (correct refusals of harmful queries). Our work uniquely combines adversarial generation with false positive minimization: we train a proxy model to simulate refusal behavior, iteratively refine it via active learning, and generate adversarial benign prompts using explainable triggers—forming a closed-loop system for evaluation and data augmentation. This integration of dynamic testing, attribution, and iterative proxy training distinguishes our approach from prior art.

\end{document}